\documentclass{sig-alternate-05-2015}

\usepackage{comment}
\usepackage{amsmath,bm}
\usepackage{amsfonts}
\usepackage{xcolor}
\usepackage{multirow}
\usepackage{tabularx}
\usepackage{flushend}
\usepackage{booktabs}

\begin{document}

\setcopyright{acmcopyright}

\newcommand\Tstrut{\rule{0pt}{2.7ex}}         
\newcommand\Bstrut{\rule[-0.9ex]{0pt}{0pt}}   

\newcommand{\xavi}[1]{\textcolor{red}{\textbf{X:} #1}}
\newcommand{\glo}[1]{\textcolor{blue}{\textbf{G:} #1}}
\newcommand{\eli}[1]{\textcolor{magenta}{\textbf{E:} #1}}

\doi{10.475/123_4}

\isbn{123-4567-24-567/08/06}


\acmPrice{\$15.00}

%

\title{How to Make an Image More Memorable? \\A Deep Style Transfer Approach}

%
%
%
%
%

\numberofauthors{1}
\author{
Aliaksandr Siarohin$^1$, Gloria Zen$^1$, Cveta Majtanovic$^1$,\\ Xavier Alameda-Pineda$^2$, Elisa Ricci$^{3}$ and Nicu Sebe$^1$\\
\affaddr{$^1$University of Trento, \texttt{name.lastname@unitn.it}}\\
\affaddr{$^2$INRIA Grenoble, \texttt{xavier.alameda-pineda@inria.fr}}\\
\affaddr{$^{3}$Fondazione Bruno Kessler and University of Perugia, \texttt{eliricci@fbk.eu}}
}

\maketitle
\begin{abstract}
Recent works have shown that it is possible to automatically predict intrinsic image properties like memorability. In this paper, we take a step forward addressing the question: ``\emph{Can we make an image more memorable?}''.
Methods for automatically increasing image memorability would have an impact in many application fields like education, gaming or advertising.
Our work is inspired by the popular \textit{editing-by-applying-filters} paradigm adopted in photo editing applications, like Instagram and Prisma.
In this context, the problem of increasing image memorability maps to that of retrieving ``memorabilizing'' filters or style ``seeds''.
Still, users generally have to go through most of the available filters before finding the desired solution, thus turning the editing process into a resource and time consuming task. 
%
In this work, we show that it is possible to automatically retrieve the \emph{best} style seeds for a given image, thus 
remarkably reducing the number of human attempts needed to find a good match. 
Our approach leverages from recent advances in the field of image synthesis and adopts a deep architecture for generating a memorable picture from a given input image and a style seed.  Importantly, to automatically select the best style a novel learning-based solution, also relying on deep models, is proposed.
Our experimental evaluation, conducted on publicly available benchmarks, demonstrates the effectiveness of the proposed approach for generating memorable images through automatic style seed selection. 
\end{abstract}

%
%

\ccsdesc[500]{Computing methodologies~Computer vision}
\ccsdesc[300]{Computing methodologies~Computational photography}

%
%

%
%
\printccsdesc


\keywords{Memorability; photo enhancement; deep style transfer; retrieval.}

\begin{figure}[t]
\includegraphics[width=\linewidth]{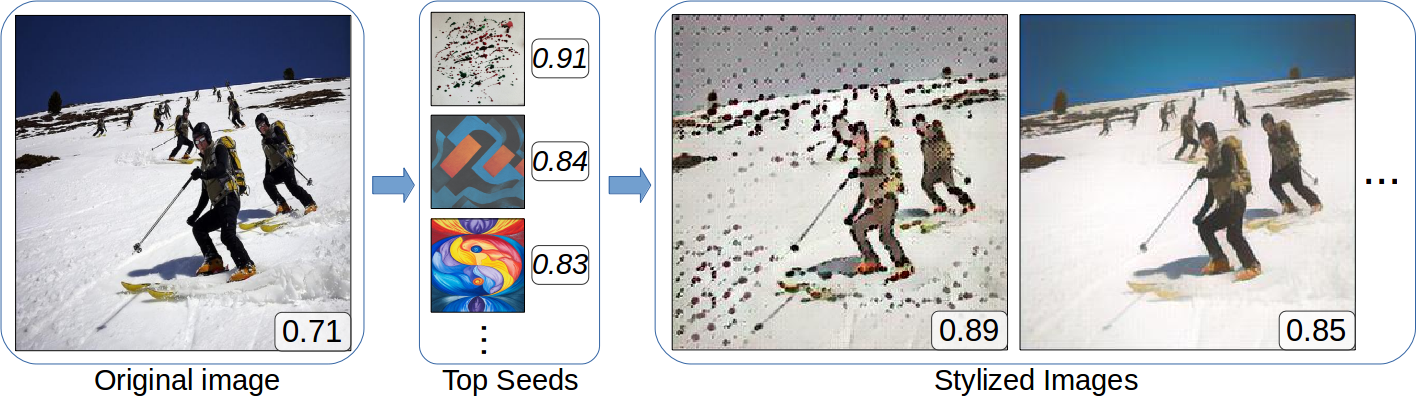}
\vspace{-0.7cm}
\caption{Sample results illustrating our idea (best viewed in colors). Given a generic image (left), automatically find the best style filters (center) that augment the memorability of the image (right). Memorability scores in the range [0,1] are reported in the small boxes.}
\label{fig:teaser}
\vspace{-0.3cm}
\end{figure}

\vspace{1.0cm}
\section{Introduction}

Today's expansion of infographics is certainly related to one of the everyday life idiom ``A picture is worth a thousand words'' (or more precisely 84.1 words~\cite{blackwell1997correction}) and to the need of providing the fastest possible knowledge transfer in the current ``information overload'' age. A recent study~\cite{hilbert2012much} showed that everyone is bombarded by the equivalent of 174 newspapers of data every day. In this context, we ask ourselves: \emph{Is it possible to transform a user-chosen image so that it has more chances to be remembered}?


For this question to be properly stated, it requires the existence of a measure of memorability, and recent studies proved that memorability is intrinsic to the visual content and is measurable~\cite{isola2011what,isola2014what}. Indeed, these studies use the \textit{memory pairs game} to provide an objective evaluation of image memorability, which has surprisingly low variance across trials.
Recent studies have also provided tools to detect the visual features responsible for both memorable and easily forgettable images. For instance, images that tend to be forgotten lack distinctiveness, like natural landscapes, whereas pictures with people, specific actions and events or central objects are way more memorable~\cite{brady2008visual}. 
Previous papers have also analyzed the relationship between emotions and memorability~\cite{cahill1995novel}. 
In a similar line of though, researchers wondered how to accurately predict which images will be remembered and which will be not. Recent experiments showed near-human performance in estimating, measuring and predicting visual memorability~\cite{khosla2015understanding}, where MemNet, a model trained on the largest annotated image memorability dataset, LaMem, has been proposed.

While previous studies on automatic prediction of memorability from images paved the way towards the automatic recognition of image memorability, many questions are still open. For instance: \textit{is it possible to increase the memorability of an image, while keeping its high-level content}? Imagine an advertising campaign concerning the design for a new product targeting a specific market sector. Once the very expensive designing phase is over, the client receives a set of images advertising the new product. Such images tell a story: in the attempt of increasing the image's memorability, the high-level content, that is \textit{the meaning}, should remain intact. We therefore focus on how to automatically modify the style of the image, that is how to \textit{filter} the image, so as to make it more memorable.

Some popular commercial products are based on this image filtering philosophy, for other purposes than memorability though. For instance, Instagram\footnote{https://www.instagram.com}, a photo and video sharing Internet platform launched in 2010, allows the users to filter the visual content with several pre-defined filters before sharing. 
Similarly, Prisma\footnote{http://prisma-ai.com} 
turns user memories into art by using artificial intelligence. 
In parallel to the development of these commercial products, several recent research studies in computer vision and multimedia have focused on creating artistic images of high perceptual quality with artificial intelligence models. For instance, Gatys \textit{et al.}~\cite{gatys2016image} have proposed an approach where a deep network is used to manipulate the content of a natural image adapting it to the style of a given artwork. Subsequently, more efficient deep architectures for implementing a style transfer have been introduced~\cite{ulyanov2016texture}. Importantly, none of these commercial products and related research studies incorporate the notion of image memorability. 

In this work, we propose a novel approach for increasing the memorability of images which is inspired by the \textit{editing-by-filtering} framework (Fig.~\ref{fig:teaser}). Our method relies on three deep networks. A first deep architecture, the \textit{Synthesizer} network, is used to synthesize a memorable image from the input picture and a style picture. A second network acts as a style \textit{Selector} and it is used to retrieve the ``best'' style seed to provide to the Synthesizer, (\textit{i.e.} the one that will produce the highest increase in terms of memorability) given the input picture. 
To train the Selector, pairs of images and vectors of memorability gap scores (indicating the increase/decrease in memorability when applying each seed to the image) are used. A third network, the \textit{Scorer}, which predicts the memorability score from a given input image, is used to compute the memorability gaps necessary to train the Selector.
Our approach is extensively evaluated on the publicly available LaMem dataset \cite{khosla2015understanding} and we show that it can be successfully used to automatically increase the memorability of natural images.


The main contributions of our work are the following:
\vspace{-0.1cm}
\begin{itemize}
\setlength\itemsep{0.0cm}
    \item We tackle the challenging task of increasing image memorability while keeping the high-level content intact (thus modifying only the style of the image).
    \item We cast this into a style-based image synthesis problem using deep architectures 
    and propose an automatic method to retrieve the style seeds that are expected to lead to the largest increase of memorability for the input image.
    \item We propose a lightweight solution for training the Selector network implementing the style seed selection process, allowing us to efficiently learn our model with a reduced number of training data while considering  relatively large variations of style pictures.
\end{itemize}



\begin{figure*}[t]
\includegraphics[width=\linewidth]{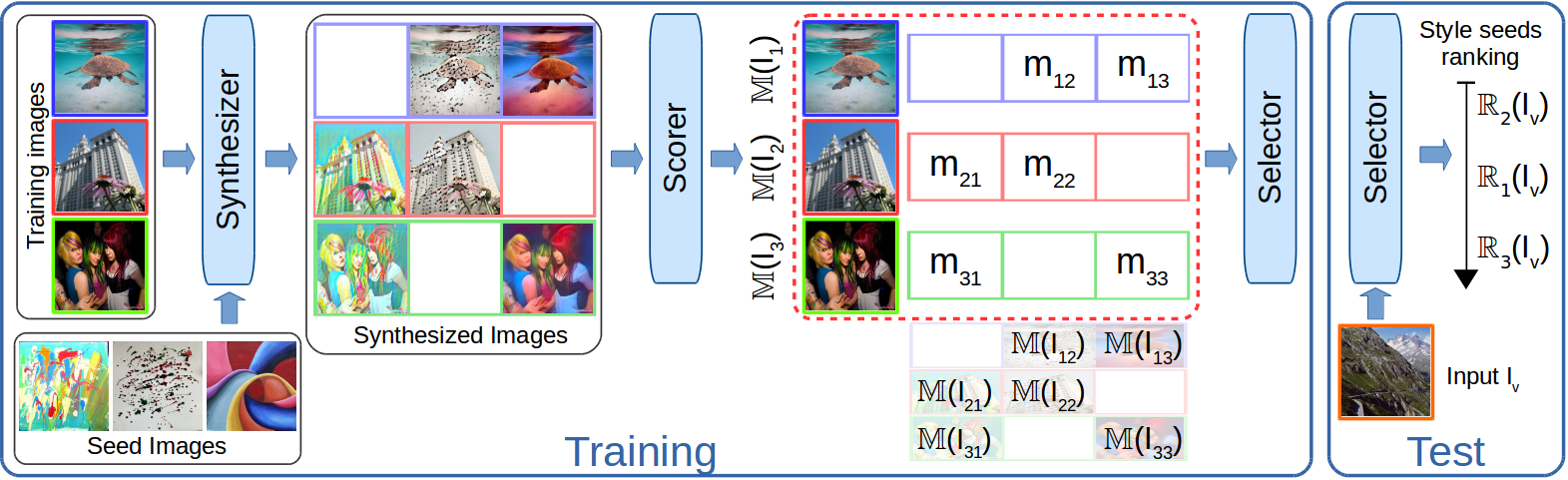}
\vspace{-0.6cm}
\caption{Overview of our method. At training time, the Synthesizer $\mathbb{S}$ and the Scorer $\mathbb{M}$ serve to generate the training data (highlighted with a red dotted frame) for the seed Selector $\mathbb{R}$. 
At test time, the seed Selector provides for each new image a sorted list of style seeds, based on the predicted memorability increase $\mathbb{R}_{s}(\mathbf{I}_v)$.  }
\label{fig:method}
\vspace{-0.3cm}
\end{figure*}

\section{Related Works}
The concept of memorability and its relation with other aspects of the human mind has been long studied from a psychological perspective~\cite{maren1999long,anderson2006emotion,brady2008visual,hunt2006distinctiveness,phelps2004human,bradley1992remembering}. 
Works in psychology and neuroscience mostly focused on visual memory, studying for instance the human capacity of remembering object details~\cite{brady2008visual}, the effect of emotions on memory~\cite{anderson2006emotion,phelps2004human,bradley1992remembering} or the brain's learning mechanisms, \textit{e.g.} the role of the amygdala in memory~\cite{maren1999long,phelps2004human}.
For a few years now, more automated studies on memorability have arisen: from the collection of image datasets specifically designed to study memorability, to user-friendly techniques to annotate these data with memorability scores. The community is now paying attention to understand the causes of visual memorability and its prominent links with, for instance, image content, low- and mid-level visual features and evoked emotions.

Isola \textit{et al.}~\cite{isola2011what} showed that visual memorability is an intrinsic property of images, and that it can be explained by considering only image visual features. Besides the expected inter-subject variability, \cite{isola2011what} reported a large consistency among viewers when measuring the memorability of several images. Typically, such measures are obtained by playing a memory game. Other studies proved that memorability can also be automatically predicted. Recently, Khosla \textit{et al.}~\cite{khosla2015understanding} used CNN-based features from MemNet to achieve a prediction accuracy very close to human performance, \textit{i.e.} up to the limit of the inter-subject variability, thus outperforming previous works using hand-crafted features such as objects or visual attributes~\cite{isola2014what}. 

In parallel, large research efforts have been invested in understanding what makes an image memorable and, in a complementary manner, which is the relation between image memorability and other subjective properties of visual data, such as interestingness, aesthetics or evoked emotions. Gygli \textit{et al.}~\cite{gygli2013interestingness} observed that memorability negatively correlates with visual interestingness. Curiously, they also showed that human beings perform quite bad at judging the memorability of an image, thus further justifying the use of memory games for annotation. In the same study, it was shown that aesthetic, visual interestingness and human judgements of memorability are highly correlated. Similar results were reported later on in~\cite{isola2014what}, confirming these findings. A possible mundane interpretation of these findings is that people wish to remember what they like or find interesting, though this is not always the case.

Khosla \textit{et al.}~\cite{khosla2015understanding} showed that, with the exception of \emph{amusement}, images that evoke negative emotions like \emph{disgust}, \emph{anger} and \emph{fear} are more likely to be remembered. Conversely, images that evoke emotions like \emph{awe} and \emph{contentment} tend to be less memorable. Similarly, the authors of~\cite{isola2011understanding} showed that attributes like \emph{peaceful} are negatively correlated with memorability. Other works showed that \emph{arousal} has a strong effect on human memory~\cite{anderson2006emotion,cahill1995novel,bradley1992remembering,mcgaugh2006make} at two different stages: either during the encoding of visual information (\textit{e.g.}, increased attention and/or processing) or post-encoding (\textit{e.g.}, enhanced consolidation when recalling the stored visual information). Memorability was also investigated with respect to distinctiveness and low-level cues such as colors in~\cite{borkin2013makes} and with respect to eye fixation in~\cite{khosla2015understanding,bylinskii2015intrinsic}. In more detail, \cite{borkin2013makes} discussed how images that stand out of the context (\textit{i.e.}, they are unexpected or unique) are more easily remembered and that memorability significantly depends upon the number of distinct colors in the image. 
These findings support our intuition that it is possible to {manipulate an image to increase its memorability}.
Indeed, this can happen for example by indirectly modifying image distinctiveness or the evoked arousal. Along the same line of though, Peng \textit{et al.} \cite{peng2015mixed} attempted to modify the emotions evoked by an image adjusting its color tone and its texture-related features.

Recent works analyzed how images can be modified to increase or decrease their memorability~\cite{khosla2013modifying,khosla2015understanding}. These are based on other contemporary studies that focused on generating memorability maps of images~\cite{khosla2012image,khosla2012memorability,khosla2014what}. In particular, Khosla \textit{et al.}~\cite{khosla2015understanding} showed that by removing visual details from an image through a cartonization process the memorability score can be modified. However, they did not provide a methodology to systematically increase the memorability of pictures. The same group~\cite{khosla2013modifying} also demonstrated that it is possible to increase the memorability of faces, while maintaining the identity of the person and properties like age, attractiveness and emotional magnitude. Up to our knowledge, this is the first attempt to automatically increase the memorability of generic images (not only faces).

\section{Method}

In this section we introduce the proposed framework to automatically increase the memorability of an input image. Our method is designed in a way such that the process of ``memorabilizing'' images is performed in an efficient manner while preserving most of the high-level image content. 

\subsection{Overview}
The proposed approach co-articulates three main components, namely: the seed Selector, the Scorer and the Synthesizer, and so we refer to it as $S^3$ or \emph{S-cube}. In order to give a general idea of the overall methodological framework, we illustrate the pipeline associated to S-cube in Figure~\ref{fig:method}. The {Selector} is the core of our approach: for a generic input image $\textbf{I}$ and given a set of style image seeds ${\cal S}$, the Selector retrieves the subset of ${\cal S}$ that will be able to produce the largest increase of memorability. In details, the seed Selector predicts the expected increase/decrease of memorability that each seed $\mathbf{S}\in{\cal S}$ will produce in the input image $\mathbf{I}$, and consequently it ranks the seeds according to the expected increase of memorability. At training time, the Synthesizer and the Scorer are used to generate images from many input image-seed pairs and to score these pairs, respectively. Each input image is then associated to the relative increase/decrease of memorability obtained with each of the seeds. With this information, we can learn to predict the increase/decrease of memorability for a new image, and therefore rank the seeds according to the expected increase. Indeed, at query time, the Selector is able to retrieve the most memorabilizing seeds and give them to the Synthesizer. In the following, we first formalize the S-cube framework and then describe each of the three components in detail.


\subsection{The S-cube approach}
\label{sec:scube}
Let us denote the Scorer, the Synthesizer and the seed Selector models by $\mathbb{M}$,  $\mathbb{S}$ and $\mathbb{R}$, respectively. During the \textbf{training phase} the three models are learned.
The Scoring model $\mathbb{M}$ returns the memorability value of a generic image $\mathbf{I}$, $\mathbb{M}(\mathbf{I})$, and it is learned by means of a training set of images annotated with memorability: ${\cal M}=\{\mathbf{I}_i^{\cal M},m_i\}_{i=1}^I$. In addition to this training set, we also consider a generating set of natural images ${\cal G}=\{\mathbf{I}_g^{\cal G}\}_{g=1}^G$ and a set of style seed images ${\cal S}=\{\mathbf{S}_s\}_{s=1}^S$. The Synthesizer produces an image from an image-seed pair, 
\begin{equation}
    \mathbf{I}_{gs} = \mathbb{S}\left(\mathbf{I}_g^{\cal G},\mathbf{S}_s\right).
\end{equation}\\
The scoring model $\mathbb{M}$ and the Synthesizer $\mathbb{S}$ are the required steps to train the seed Selector $\mathbb{R}$. Indeed, for each  image $\mathbf{I}_g^{\cal G}\in{\cal G}$ and for each style seed $\mathbf{S}_s\in{\cal S}$, the synthesis procedure generates $\mathbf{I}_{gs}$.
The Scoring model is used to compute the memorability score gap between the synthesized and the original images:
\begin{equation}
m_{gs}^\mathbb{M} = \mathbb{M} (\mathbf{I}_{gs}) - \mathbb{M} (\mathbf{I}_g^{\cal G}).
\end{equation}
The seed-wise concatenation of these scores, denoted by $\mathbf{m}_g^\mathbb{M} = (m_{gs}^\mathbb{M})_{s=1}^S$, is used to learn the seed Selector. Specifically, a training set of natural images labeled with the seed-wise concatenation of memorability gaps ${\cal R}=\{\mathbf{I}_g^{\cal G},\mathbf{m}_g^\mathbb{M}\}_{g=1}^{G}$ is constructed. The process of seed selection is casted as a regression problem and the mapping $\mathbb{R}$ between an image and the associated vector of memorability gap scores is learned. This indirectly produces a ranking of the seeds in terms of their the ability to memorabilize images (\textit{i.e.} the best seed corresponds to the largest memorability increase).

During the \textbf{test phase} and given a novel image $\mathbf{I}_v$, the seed Selector is applied to predict the vector of memorability gap scores associated to all style seeds, \textit{i.e.} $\mathbf{m}_v=\mathbb{R}(\mathbf{I}_v)$. A ranking of seeds is then derived from the vector $\mathbf{m}_v$. Based on this ranking the Synthesizer is applied to the test image $\textbf{I}_v$ considering only the top $Q$ style seeds $\textbf{S}_s$ and produces a set of stylized images $\{\mathbf{I}_{qs}\}_{q=1}^Q$. 

In the following we describe the three main building blocks of our approach, providing details of our implementation.

\subsection{The Scorer}
\label{sec:scoring}

\begin{sloppypar}
The scoring model $\mathbb{M}$ returns an estimate of the memorability associated to an input image $\textbf{I}$. In our work, we use the memorability predictor based on LaMem in~\cite{khosla2015understanding}, which is the state of the art to automatically compute image memorability. In details, following~\cite{khosla2015understanding} we consider a hybrid CNN model~\cite{hybrydcnn}. The network is pre-trained first for the object classification task (\textit{i.e.} on ImageNet database) and then for the scene classification task (\textit{i.e.} on Places dataset). Then, we randomly split the LaMem {training set}
into 
two disjoint subsets (of 22,500 images each), ${\cal M}$ and ${\cal E}$. We use the pretrained model and the two subsets to learn two independent scoring models $\mathbb{M}$ and $\mathbb{E}$. While, as discussed above, $\mathbb{M}$ is used during the training phase of our approach, the model $\mathbb{E}$ is adopted for evaluation (see Section \ref{sec:dataset}). For training, we run $70$k iterations of stochastic gradient descent with momentum 0.9, learning rate $10^{-3}$ and batch size 256. 
\end{sloppypar}

\begin{figure}[t]
\centering
\begin{tabular}{ccccc}
\hspace{-0.3cm}
\frame{\includegraphics[height=1.5cm, width=1.5cm]{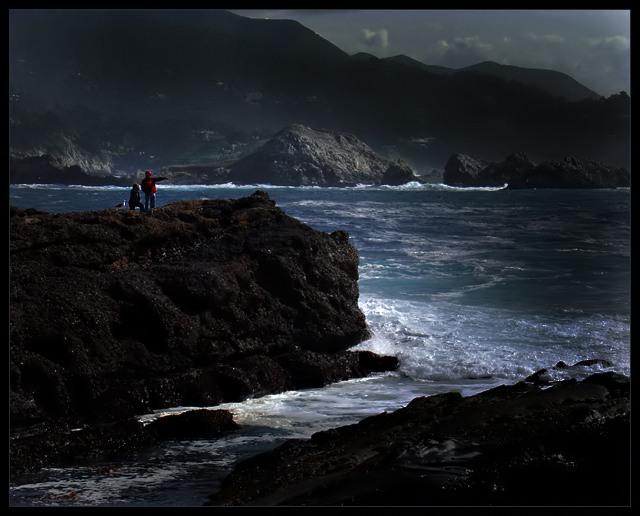}} &
\hspace{-0.3cm}
\frame{\includegraphics[height=1.5cm, width=1.5cm]{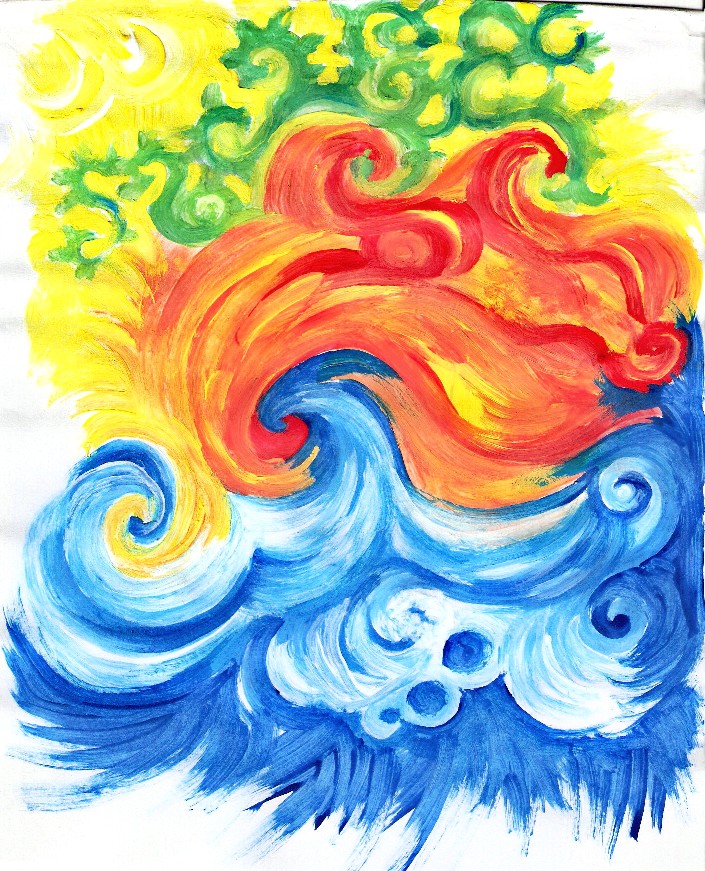}} &
\hspace{-0.1cm}
\frame{\includegraphics[height=1.5cm, width=1.5cm]{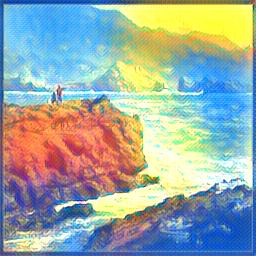}} &
\hspace{-0.3cm}
\frame{\includegraphics[height=1.5cm, width=1.5cm]{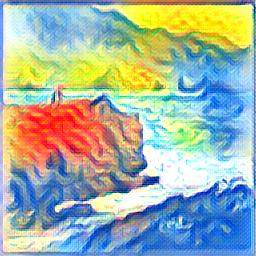}} &
\hspace{-0.3cm}
\frame{\includegraphics[height=1.5cm, width=1.5cm]{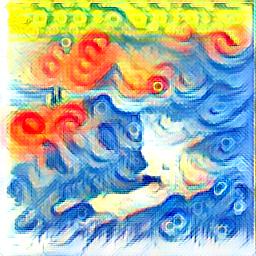}} \\
\hspace{-0.3cm}
\frame{\includegraphics[height=1.5cm, width=1.5cm]{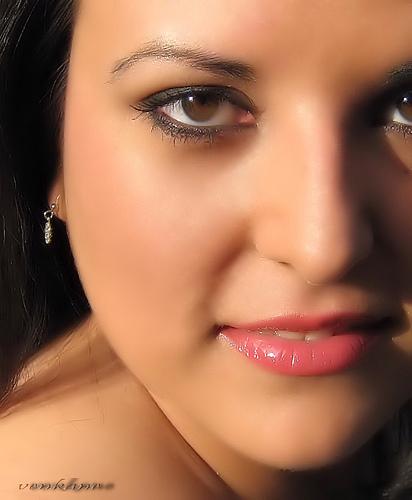}} &
\hspace{-0.3cm}
\frame{\includegraphics[height=1.5cm, width=1.5cm]{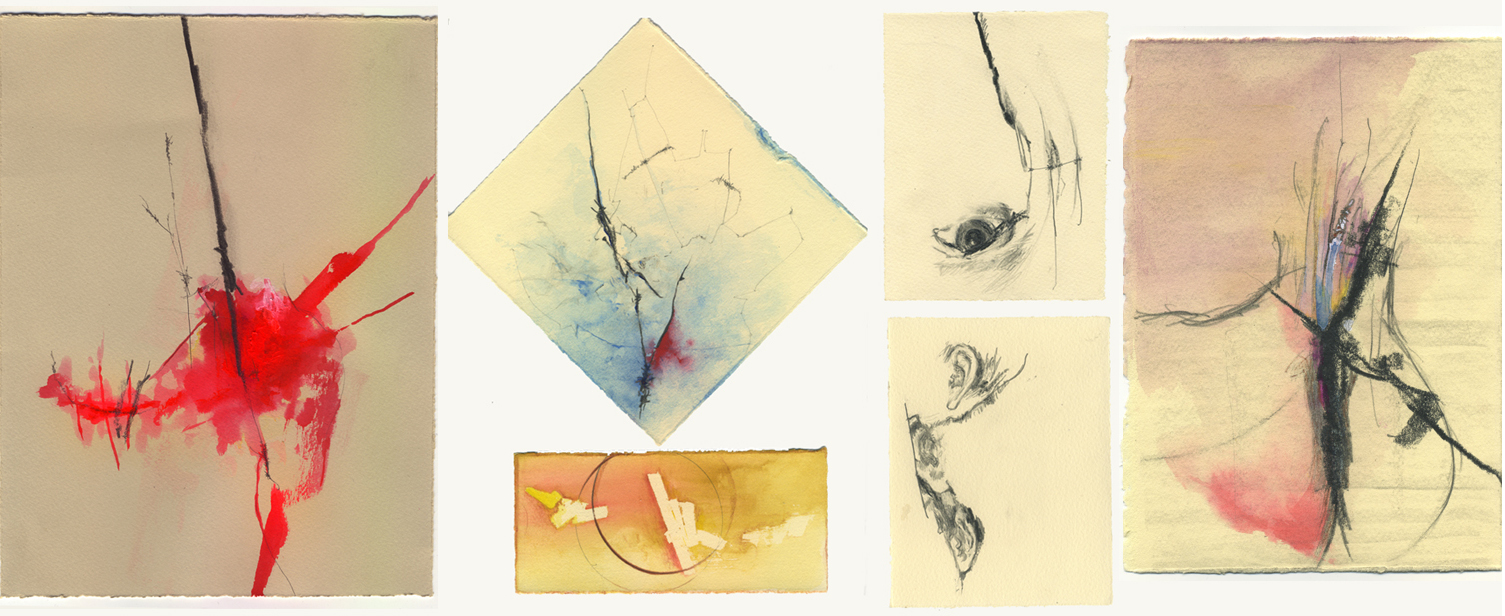}} &
\hspace{-0.1cm}
\frame{\includegraphics[height=1.5cm, width=1.5cm]{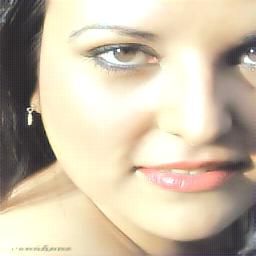}} &
\hspace{-0.3cm}
\frame{\includegraphics[height=1.5cm, width=1.5cm]{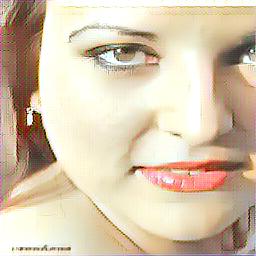}} &
\hspace{-0.3cm}
\frame{\includegraphics[height=1.5cm, width=1.5cm]{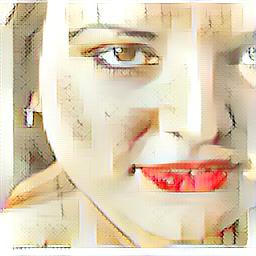}} \\
\textbf{I}$_v$ & \textbf{S}$_s$ & $0.5$ & $2$ & $10$ 
\end{tabular}
\caption{Sample results. (Left) Original images and applied style seeds. (Right) Synthesized images at varying parameter $\alpha$, which regulates the trade-off between preserving the original content of the given image $\mathbf{I}_v$ and transferring the style $\mathbf{S}_s$. 
} 
\label{fig:style}
\end{figure}

\subsection{The Synthesizer}
\label{sec:synthesis}
The {Synthesizer} takes as an input a generic image $\textbf{I}_g$ and a style seed image $\textbf{S}_s$ and produces an stylized image $\mathbf{I}_{gs} = \mathbb{S}(\textbf{I}_g,\mathbb{S}_s)$. We use the strategy proposed in~\cite{ulyanov2016texture}, which consists on training a different feed-forward network for every seed. As seeds, we use 100 abstract paintings from the DeviantArt database~\cite{sartori2015affective}, and therefore we train $S=100$ networks for $10$k iterations with learning rate $10^{-2}$. The most important hyper-parameter is the coefficient $\alpha$, which regulates the trade-off between preserving the original image content and producing something closer to the style seed (see Figure \ref{fig:style}). In our experiments we evaluated the effect of $\alpha$ (see Section \ref{sec:exp}). 
It is worth noticing that the methodology proposed in this article is independent of the synthesis procedure. Indeed, we also tried other methods, namely Gatys \textit{et\ al.}~\cite{gatys2015neural} and Li \textit{et\ al.}~\cite{li2016precomputed}, but we selected~\cite{ulyanov2016texture} since it showed very good performance while keeping low computational complexity. This is especially important in our framework since the Synthesizer is also used to generate the training set for learning $\mathbb{R}$. 


\subsection{The Seed Selector}
\label{sec:seed_select}
The core part of our approach is the Selector.
Given a training set of natural images labeled with the vector of memorability gaps: ${\cal R}=\{\mathbf{I}_g^{\cal G},\mathbf{m}_g^\mathbb{M}\}_{g=1}^{G}$, the seed Selector $\mathbb{R}$ is trained minimizing the following objective:
\begin{equation}
 {\cal L}_\mathbb{R} = \sum_{g=1}^G  \mathcal{L}\left(\mathbb{R}(\mathbf{I}_g^{\cal G}),\mathbf{m}_g^\mathbb{M}\right).
 \label{loss}
\end{equation}
where $\mathcal{L}$ is a loss function which measures the discrepancy between the learned vector $\mathbb{R}(\mathbf{I}_g^{\cal G})$ and the memorability gap scores $\mathbf{m}_g^\mathbb{M}$.
By training the seed Selector with memorability gaps, we are learning \textit{by how much each of the seeds increases or decreases the memorability of a given image}. This has several advantages. First, we can very easily rank the seeds by the expected increase in memorability they will produce if used together with the input image and the synthesis procedure. Second, if several seeds have similar expected memorability increase, they can be proposed to the user for further selection. Third, if all seeds are expected to decrease the memorability, the optimal choice of not modifying the image can easily be made. Fourth, once $\mathbb{R}$ is trained, all this information comes at the price of evaluating $\mathbb{R}$ for a new image, which is cheaper than running $\mathbb{S}$ and $\mathbb{M}$ $S$ times.

Even if this strategy has many advantages at testing time, the most prominent drawback is that, to create the training set $\mathcal{R}$, one should ideally call the synthesis procedure for all possible image-seed pairs. This clearly reduces the scalability and the flexibility of the proposed approach. The scalability because training the model on a large image dataset means generating a much larger dataset (\textit{i.e.}, $S$ times larger). The flexibility because if one wishes to add a new seed to the set ${\cal S}$, then all image-seed pairs for the new seed need to be synthesized and this takes time. Therefore, it would be desirable to find a way to overcome these limitations while keeping the advantages described in the previous paragraph.

The solution to these issues comes with a model able to learn from a partially synthesized set, in which not all image-seed pairs are generated and scored. This means that the memorability gap vector $\mathbf{m}^{\mathbb M}_g$ has  missing entries. In this way we only require to generate \textit{enough} image-seed pairs. To this aim, we propose to use a decomposable loss function $\mathcal{L}$. Formally, we define a binary variable $\omega_{gs}$ set to $1$ if the 
$gs$-th image-seed pair is available and to $0$ otherwise and rewrite the objective function in (\ref{loss}) as:
\begin{equation}
 {\cal L}_\mathbb{R} = \sum_{g=1}^G \sum_{s=1}^S \omega_{gs}\ell\left(\mathbb{R}_s(\mathbf{I}_g^{\cal G}),m_{gs}^\mathbb{M}\right).
\end{equation}
where $\mathbb{R}_s$ is the $s$-th component of $\mathbb{R}$ and $\ell$ is the square loss.
We implemented this model using an AlexNet architecture, where the prediction errors for the missing entries of $\mathbf{m}^{\mathbb{M}}_g$ are not back-propagated. 
Specifically, we considered the pre-trained Hybryd-CNN and fine-tune only the layers fc6, fc7, conv5, conv4 using learning rate equal to $10^{-3}$, momentum equal to 0.9 and batch size 64. The choice of  Hybryd-CNN is considered more appropriate when dealing with generic images since the network is pre-trained both on images of places and objects.

\begin{table*}[t]
\begin{center}
\scalebox{0.95}{
\begin{tabular*}{0.9\linewidth}{cc@{\extracolsep{\fill}}c cc cc c cc cc}
\toprule
& & & \multicolumn{2}{c}{$A^{\mathbb M}$} &  \multicolumn{2}{c}{$A^{\mathbb E}$} && \multicolumn{2}{c}{MSE$^{\mathbb M}$} &  
\multicolumn{2}{c}{MSE$^{\mathbb E}$} \Tstrut\Bstrut \\ \midrule
 $\alpha$ & $\bar{\omega}$ && $\mathcal{B}$ & S-cube & $\mathcal{B}$ & S-cube  && $\mathcal{B}$ & S-cube & $\mathcal{B}$ & S-cube  
\Tstrut\Bstrut  \\  \midrule
\multirow{4}{*}{2} & 0.01 && \textbf{63.21} &	57.12 & \textbf{60.96}	& 56.01  && \textbf{0.0113} & 0.0138 & \textbf{0.0119} & 0.0137 \Tstrut  \\
 & 0.1  && 64.49 & \textbf{64.70} & 61.07	& \textbf{62.22} && \textbf{0.0112} & 0.0114 & \textbf{0.0117} & 0.0119  \\
 & 0.5  && 64.41 & \textbf{67.18} & 61.06	& \textbf{64.38} && 0.0112 &	\textbf{0.0102} & 0.0117 & \textbf{0.0106} \\
 & 1    && 64.41 & \textbf{67.80} & 61.06	& \textbf{64.71} && 0.0112 &	\textbf{0.0102} & 0.0117 & \textbf{0.0108} \\
\midrule
\multirow{4}{*}{10} & 0.01 && \textbf{67.91}  & 64.74 & \textbf{68.31} & 64.74 && \textbf{0.0126} & 0.0151 & \textbf{0.0134} & 0.0163 \Tstrut \\
 & 0.1  && 68.04  & \textbf{72.25} & 68.36 & \textbf{70.96} && 0.0125 & \textbf{0.0116} & 0.0132 & \textbf{0.0121} \\
 & 0.5  && 67.99  & \textbf{73.26} & 68.31 & \textbf{71.72} && 0.0125 & \textbf{0.0109} & 0.0132 & \textbf{0.0112} \\
 & 1    && 68.04  & \textbf{73.26} & 68.31 & \textbf{71.75} && 0.0125 & \textbf{0.0108} & 0.0132 & \textbf{0.0111} \\\bottomrule
\end{tabular*}
}
\vspace{-0.2cm}
\end{center}
\caption{Performance of our method S-cube compared to baseline $\mathcal{B}$ at varying percentage of training data $\bar{\omega}$ and style 
coefficient $\alpha$, measured in terms of (left) accuracy $A$ and (right) mean squared error (MSE). Performances have been evaluated using both the 
internal $\mathbb{M}$ and external $\mathbb{E}$ predictor.}
\label{tab:perf}
\end{table*}

\section{Experimental Validation}
\label{sec:exp}
\begin{sloppypar}
We assess the performance of our approach in successfully retrieving the most memorabilizing seeds to increase the memorability of arbitrary images (Sec.~\ref{sec:res-memorability}).
The datasets and experimental protocol used in our study are described in Sec.~\ref{sec:dataset}.

\subsection{Datasets and Experimental Protocol}
\label{sec:dataset}
In our experiments we consider two publicly available datasets, LaMem\footnote{http://memorability.csail.mit.edu} and DeviantArt\footnote{http://disi.unitn.it/$\sim$sartori/datasets/deviantart-dataset}.

\textbf{LaMem.} The LaMem dataset \cite{khosla2015understanding} is the largest dataset used to study memorability. It is a collection of 58,741 images gathered from a number of previously existing datasets, including the affective images dataset~\cite{machajdik2010affective}, which consists of Art and Abstract paintings. The memorability scores were collected for all the dataset pictures using an optimized protocol of the memorability game. The corpus was released to overcome the limitations of previous works on memorability which considered small datasets and very specific image domains. The large appearance variations of the images makes LaMem particularly suitable for our purpose.  

\textbf{DeviantArt.} This dataset~\cite{sartori2015affective} consists of a set of 500 abstract art paintings collected from deviantArt (dA), an online social network site devoted to user-generated art. Since the scope of our study requires avoiding substantial modifications of the high-level content of the image, we selected the style seeds from abstract paintings. Indeed, abstract art relies in textures and color combinations, thus an excellent candidate when attempting the automatic modification of the low-level image content.

\textbf{Protocol.} In our experiments using the LaMem dataset we consider the same training (45,000 images), test (10,000 images) and validation (3,741 images) data adopted in~\cite{khosla2015understanding}. We split the LaMem training set into two subsets of 22,500 images each (see also Section \ref{sec:scoring}), ${\cal M}$ and ${\cal E}$, which are used to train two predictors $\mathbb{M}$ and 
$\mathbb{E}$, respectively. The model $\mathbb{M}$ is the Scorer employed in our framework, while $\mathbb{E}$ (which we will denote in the following as the external predictor) is used to evaluate the performance of our approach, as a proxy for human assessment. We highlight that $\mathbb{M}$ and $\mathbb{E}$ can be used as two independent memorability scoring functions, since ${\cal M}$ and ${\cal E}$ are disjoint. The validation set is used to implement the early stopping.
To evaluate the performance of our scorer models ${\cal M}$ and ${\cal E}$, following \cite{khosla2015understanding}, we compute the rank correlation between predicted and actual memorability on LaMem test set. We obtain a rank correlation of 0.63 with both models, while ~\cite{khosla2015understanding} achieves a rank correlation of 0.64 training on the whole LaMem training set. As reported in ~\cite{khosla2015understanding}, this is close to human performance (0.68).

The test set of LaMem ($10$k images) is then used (i) to learn the proposed seed Selector and (ii) to evaluate the overall framework (and the Selector in particular). In detail, we split LaMem test set into train, validation and test for our Selector with proportion 8:1:1, meaning 8,000 for training and 1,000 for validation and test. The training set for the Selector was already introduced as ${\cal G}$. We denote the test set as ${\cal V}$. The validation set is used to perform early stopping, if required. 

Regarding the seeds, we estimated the memorability of all paintings of DeviantArt using $\mathbb{M}$ and selected the 50 most and he 50 least memorable images as seeds for our study (${\cal S}$). The memorability scores of the deviantArt images range from 0.556 to 0.938. 
\end{sloppypar}


\textbf{Baseline.} To the best of our knowledge this is the first work showing that it is possible to automatically increase the memorability of a generic image. For this reason, a direct and quantitative comparison with previous studies is not possible. Indeed, the recent work~\cite{khosla2015understanding} showed that it is possible to compute accurate memorability maps from images, which can be used as bases for further image manipulations. They also observed that using a memorability map for removing image details, such as through a cartoonization process, typically lead to a memorability decrease. Oppositely, we aim to effectively increase image memorability without modifying the high level content of the images. Therefore, the approach by~\cite{khosla2015understanding} does not directly compare with ours. The only potential competitor to our approach would be~\cite{khosla2013modifying}, except that the method is specifically designed for face photographs. Indeed, the proposed approach aims to modify the memorability while keeping other attributes (age, gender, expression) as well as the identify untouched. Therefore, the principle of~\cite{khosla2013modifying} cannot be straightforwardly transferred to generic images. Consequently, we define an \textit{average} baseline ${\cal B}$ that consists on ranking the style seeds according to the average memorability increase, formulated as:
\begin{equation}
\bar{m}_s^\mathbb{M} = \frac{1}{G} \sum_{g=1}^{G} m_{gs}^\mathbb{M}.
\end{equation}

\subsection{Increasing image memorability}
\label{sec:res-memorability}

We first evaluate the performance of our method at predicting the memorability increase of an image-seed match, where the seed is taken from the 
set of style seeds ${\cal S}$, and the generic image $\mathbf{I}_v^{\cal V}$ is taken from a set of (yet) unseen images ${\cal 
V}=\{\mathbf{I}_v^{\cal V}\}_{v=1}^V$. We use two different performance measures: the mean squared error (MSE) and the accuracy $A$, which are 
defined as follows:
\begin{equation}
\textrm{MSE}^\mathbb{X} = \frac{1}{SV} \sum_{s=1}^{S} \sum_{v=1}^{V} \left( m^\mathbb{X}_{vs} - \mathbb{R}_s(\mathbf{I}_v^{\cal V})\right)^2
\end{equation}
and
\begin{equation}
A^\mathbb{X} = \frac{1}{SV} \sum_{s=1}^{S} \sum_{v=1}^{V} ( 1-  | H ( m_{vs}^\mathbb{X})  -  H ( \mathbb{R}_s(\bm{I}_v^{\cal V}) ) |)
\end{equation}
where $\mathbb{X}$ indicates the internal or external predictor, respectively $\mathbb{X} = \{\mathbb{M},\mathbb{E}\}$, and $H$ is the Heaviside step function.

Table~\ref{tab:perf} reports the performance of both the proposed approach (S-cube) and the baseline (${\cal B}$) under different experimental setups. Indeed, we report the accuracy (left) and the MSE (right) evaluated using the scoring model $\mathbb{M}$ and the external scoring model 
$\mathbb{E}$ (left two and right two columns of each block), for different values of $\alpha$ and the average amount of image-seed matches $\bar{\omega}$. More precisely, $\bar{\omega}=1$ means that all image-seed pairs are used, $\bar{\omega}=0.1$ means that only 10\% is used, and 
so on.

Generally speaking our method outperforms the baseline if enough image-seed pairs are available. We argue that, as it is well known, deep architectures require a \textit{sufficient} amount of data to be effective. Indeed, when $\bar{\omega}=0.01$, the network optimization procedure attemps to learn a regression from the raw image to a 100-dimensional space with, in average, only one of this dimensions propagating the 
error back to the network. Although this dimension is different for each image, we may be facing a situation in which not enough information is 
propagated back to the parameters so as to effectively learn a robust regressor. This situation is coherent when the scoring method changes from 
$\mathbb{M}$ to $\mathbb{E}$. We can clearly observe a decrease in the performance measures when using $\mathbb{E}$, as expected. Indeed, since the 
seed selector has been trained to learn the memorability gap of $\mathbb{M}$, the performance is higher when using $\mathbb{M}$ than $\mathbb{E}$.


Furthermore, we report the performance of our method using two different values of the style coefficient $\alpha=\{2,10\}$. It can be noticed that 
our method performs better in terms of MSE when $\alpha=2$, while accuracy is usually higher for $\alpha=10$. What a priori could be 
seen as a divergent behavior, can be explained by the fact that imposing a higher weight to the style produces higher memorability gaps $m_{gs}$, 
thus it may generate a higher error in the estimation. 
We interpret these  results as an indication that MSE and $A$ can be good criteria for finding the best setup in terms percentage of training data, but not necessarily to 
set other parameters.

\begin{table}[t]
\begin{center}

\scalebox{0.95}{
\begin{tabular*}{\linewidth}{c@{\extracolsep{\fill}} cc cc}
\toprule
 & \multicolumn{2}{c}{$A^{\mathbb{E}}$} &  \multicolumn{2}{c}{MSE$^{\mathbb{E}}$} \Tstrut\Bstrut \\  \midrule
$\bar{\omega}$ & VGG16& AlexNet  & VGG16 & AlexNet   \Tstrut\Bstrut \\  \midrule
0.01 & \textbf{61.56} &  56.01 & \textbf{0.0121} & 0.0137     \Tstrut\\
0.1  & \textbf{64.76} &  62.22 & \textbf{0.0109} & 0.0119  \\
0.5  & 63.49 & \textbf{64.38} & 0.0111 & \textbf{0.0106}  \\
1    & 63.44 & \textbf{64.71} & 0.0111 & \textbf{0.0108}  \\
\bottomrule
\end{tabular*}}
\vspace{-0.2cm}
\caption{Performances of our method S-cube based on AlexNet (fine-tuning Hybrid-CNN~\cite{hybrydcnn}) and VGG16 (pre-trained on ImageNet), measured in terms of MSE$^{\mathbb{E}}$ and $A^\mathbb{E}$, at varying percentage of training data $\bar{\omega}$.}
\label{tab:vgg16}
\end{center}
\vspace{-0.3cm}
\end{table}

\begin{table}[t]
\begin{center}
\scalebox{0.95}{
\begin{tabular*}{\linewidth}{c@{\extracolsep{\fill}} cc cc}
\toprule
 & \multicolumn{2}{c}{$A^{\mathbb{E}}$} &  \multicolumn{2}{c}{MSE$^{\mathbb{E}}$} \Tstrut\Bstrut \\  \midrule
$S$ & $\cal B$ & S-cube & $\cal B$ & S-cube  \Tstrut\Bstrut \\  \midrule
20 &  60.66 & \textbf{63.15} & 0.0114 & \textbf{0.0111}    \Tstrut\\
50  & 61.09  & \textbf{63.51} & 0.0116 & \textbf{0.0109}    \\
100  & 61.06 & \textbf{64.38} & 0.0117 &  \textbf{0.0106}  \\
\bottomrule
\end{tabular*}}
\vspace{-0.2cm}
\caption{Performance of our method in terms of MSE$^\mathbb{E}$ and $A^\mathbb{E}$ ($\alpha=2$ and $\bar{\omega}=0.5$) at varying the cardinality $S$ of the style seed set.}
\label{tab:size_of_S}
\end{center}
\vspace{-0.3cm}
\end{table}

We also investigated the impact of the network depth and trained a seed Selector using VGG16 instead of AlexNet. 
We fine-tuned the layers fc6, fc7, and all conv5, using Nesterov momentum with momentum 0.9, learning rate $10^{-3}$ and batch size 64.
Importantly, while AlexNet was trained as a hybrid-CNN~\cite{khosla2015understanding}, the pre-trained model for VGG16 was trained on ImageNet. We found very interesting results and report them in Table~\ref{tab:vgg16}, for $\alpha=2$. The behavior of AlexNet was already discussed in the previous paragraphs. Interestingly we observe similar trends in VGG. Indeed, when not enough training pairs are available the results are pretty unsatisfying. However, in relative terms, the results for small $\bar{\omega}$ are far better for VGG16 than for AlexNet. We attribute this to the fact that VGG16 is much larger, and therefore the amount of knowledge encoded in the pre-trained model has a stronger regularization effect in our problem than when using AlexNet. The main drawback is that, when enough data are available and since the amount of parameters in VGG16 is much larger than in AlexNet, the latest exhibits higher performance than the former. We recall that the seed Selector is trained with 8k images, and hypothesize that fine-tuning with larger datasets (something not possible if we want to use the splits provided in LaMem) will raise the performance of the VGG16-based seed Selector.


Furthermore, we studied the behavior of the framework when varying the size $S$ of the seed set. Results are shown in Table~\ref{tab:size_of_S}. 
Specifically, we select two sets of 50 and 20 seeds out of the initial 100, randomly sampling these seeds half from the 50 most and half from the 50 least memorable ones.   
In terms of accuracy, the performance of both the proposed method and the baseline remain pretty stable when decreasing the number of seeds. This behavior was also observed in Table~\ref{tab:perf}, especially for the baseline method. However, a different trend is observed for the MSE. Indeed, while the MSE of the proposed method increases when reducing the number of seeds (as expected), the opposite trend is found for the baseline method. We argue that, even if the baseline method is robust in terms of selecting the bests seeds to a decrease of the number of seeds, it does not do a good job at predicting the actual memorability increase. Instead, the proposed method is able to select the bests seeds and better measure their impact, especially when more seeds are available. This is important if the method wants to be deployed with larger seed sets. Application-wise this is quite a desirable feature since the seeds are automatically selected and hence the amount of seeds used is transparent to the user.

\begin{figure}[t]
\hspace{-0.15cm}
\includegraphics[width=0.48\linewidth]{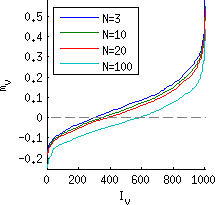}
\hspace{-0.2cm}
\includegraphics[width=0.48\linewidth]{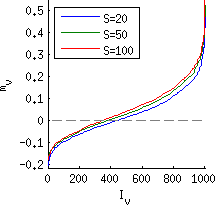}
\vspace{-0.3cm}
\caption{Sorted average memorability gaps $\bar{m}_v$ obtained with our method S-cube (left) averaging over varying number of top N seeds and (right) at varying the cardinality $S$ of the seed set, with $N=10$.}
\label{fig:delta_mem_stat1}
\vspace{-0.4cm}
\end{figure}

\begin{figure*}[t]
\begin{tabular}{ccc} 
\includegraphics[height=2.4cm, width=2.4cm]{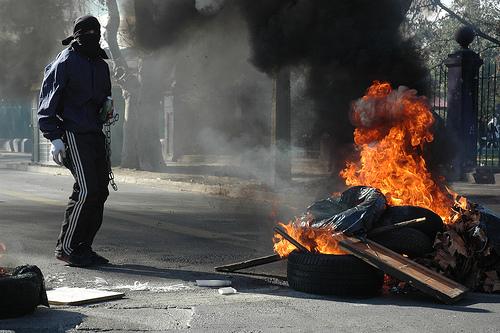} & 
\includegraphics[height=2.4cm, width=2.4cm]{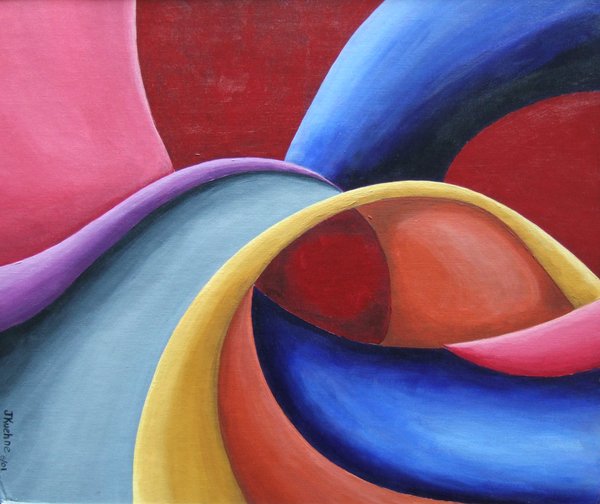} & 
\includegraphics[height=2.4cm, width=2.4cm]{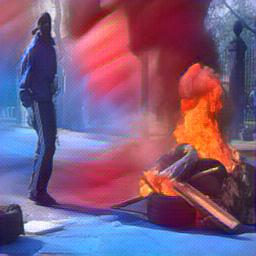} 
\\
$\mathbb{E}$: 0.72 &  & ($\mathbb{R}$: 0.76 ) $\mathbb{E}$: 0.78  \\
\includegraphics[height=2.4cm, width=2.4cm]{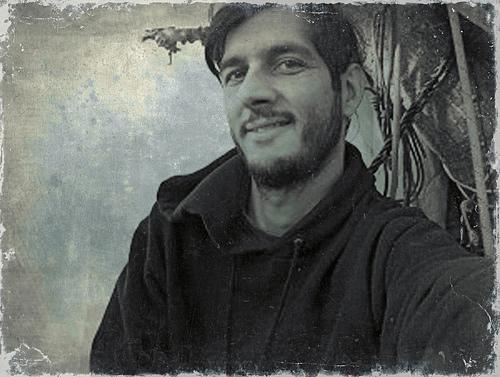} & 
\includegraphics[height=2.4cm, width=2.4cm]{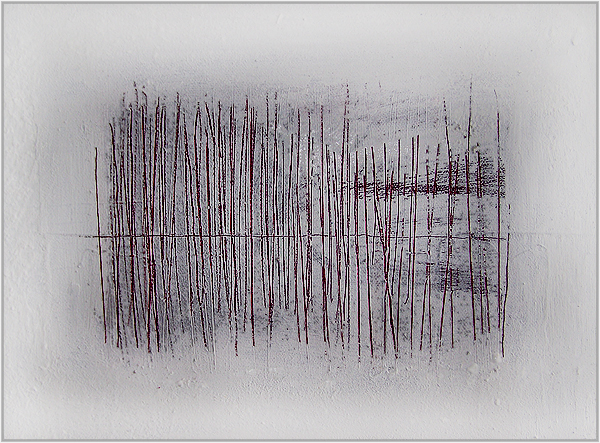} & 
\includegraphics[height=2.4cm, width=2.4cm]{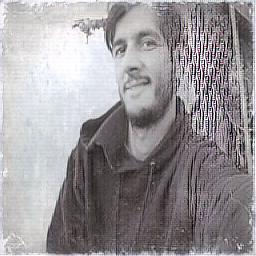} 
\\
$\mathbb{E}$: 0.78 &  & ($\mathbb{R}$: 0.78) $\mathbb{E}$: 0.85 \\
\includegraphics[height=2.4cm, width=2.4cm]{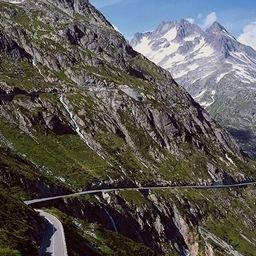} &
\includegraphics[height=2.4cm, width=2.4cm]{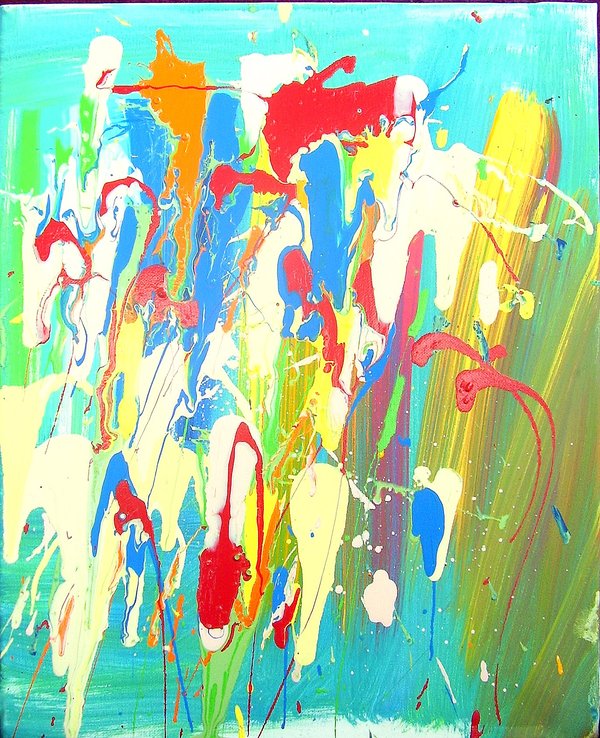} & 
\includegraphics[height=2.4cm, width=2.4cm]{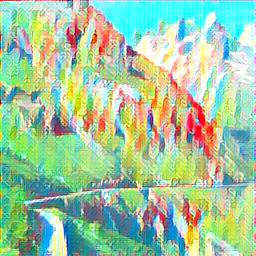} 
\\
$\mathbb{E}$: 0.54 &  & ($\mathbb{R}$:0.63) $\mathbb{E}$: 0.78 
\end{tabular}
\hspace{0.6cm}
\begin{tabular}{ccc} 
\includegraphics[height=2.4cm, width=2.4cm]{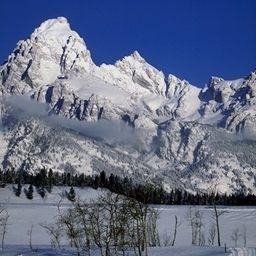} & 
\includegraphics[height=2.4cm, width=2.4cm]{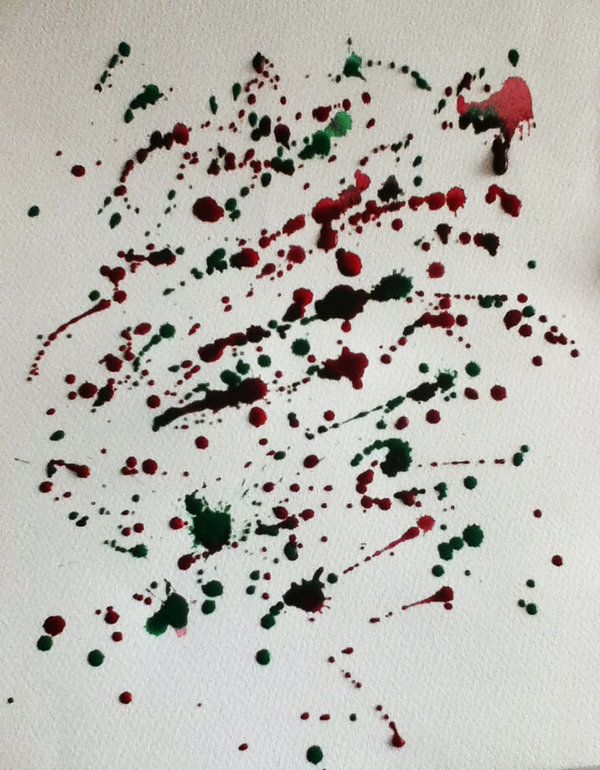} & 
\includegraphics[height=2.4cm, width=2.4cm]{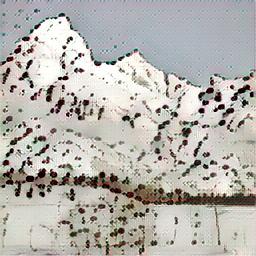} 
\\
$\mathbb{E}$: 0.60 &  & ($\mathbb{R}$: 0.78) $\mathbb{E}$: 0.81   \\
\includegraphics[height=2.4cm, width=2.4cm]{img/imgs_sw10/00041754.jpg} & 
\includegraphics[height=2.4cm, width=2.4cm]{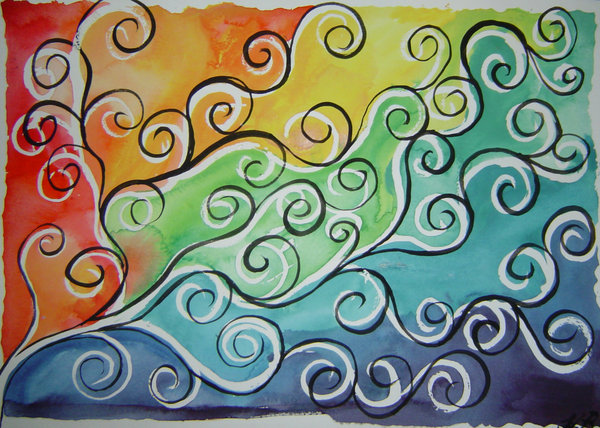} & 
\includegraphics[height=2.4cm, width=2.4cm]{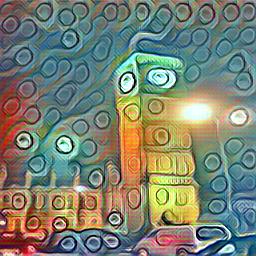} 
\\
$\mathbb{E}$: 0.67 &  & ($\mathbb{R}$: 0.80) $\mathbb{E}$: 0.82   \\
\includegraphics[height=2.4cm, width=2.4cm]{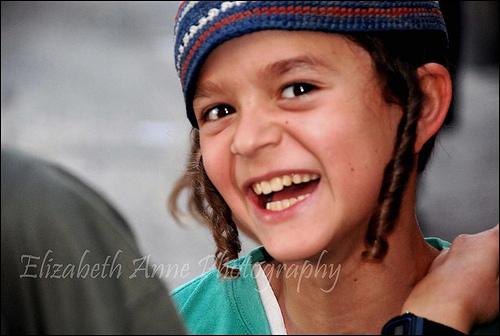} & 
\includegraphics[height=2.4cm, width=2.4cm]{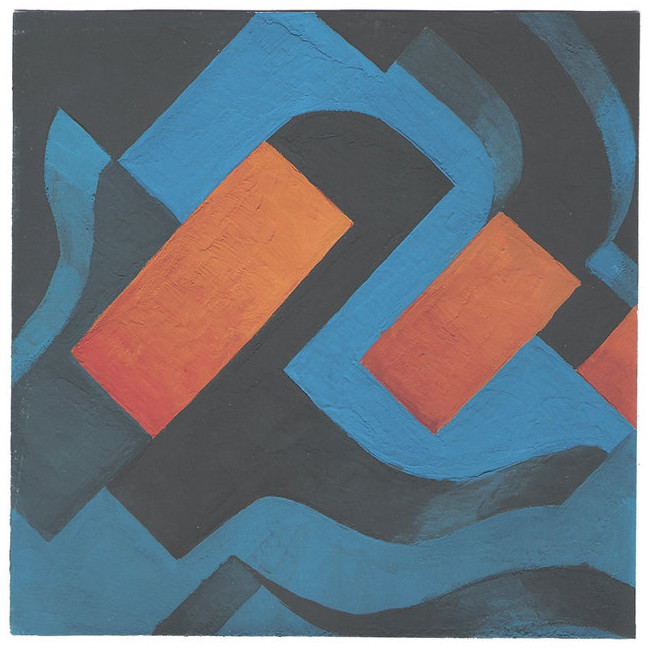} & 
\includegraphics[height=2.4cm, width=2.4cm]{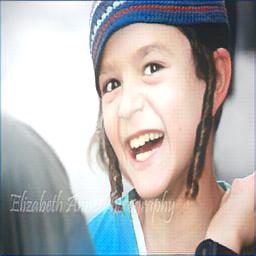} 
\\
$\mathbb{E}$: 0.72 &  & ($\mathbb{R}$: 0.69) $\mathbb{E}$: 0.89
\end{tabular}
\caption{Sample results: (left) original input image, (center) retrieved style seed and (right) corresponding synthesized image. The memorability score measured with the external model $\mathbb{E}$ is reported below each image. The memorability score predicted by the Selector $\mathbb{R}$ based on the image-seed match is reported below the resulting synthesized image.}
\label{fig:qualitative}
\end{figure*}

Finally, we assess the validity of our method as a tool for effectively increasing the memorability of a generic input image $\mathbf{I}_v$. 
In Figure~\ref{fig:delta_mem_stat1} (left) we report the average memorability gaps $\bar{m}_v$ obtained over the test set $\cal V$, when averaging over the top $N$ seeds retrieved, with $N=3,10,20$ and all the images. It can be noted that $\bar{m}_v$ achieve higher values when smaller sets of top N seeds are considered, as an indication that our method effectively retrieve the most memoralizable seeds. In Figure~\ref{fig:delta_mem_stat1} (right) we report the average memorability gaps $\bar{m}_g$ obtained over the test set $\cal V$ with our mehtod S-cube, considering $N=10$ and a varying number of style seeds $S$. It can be noted that a larger number of seeds allows to achieve higher increase.
Figure~\ref{fig:qualitative} illustrates some ``image memoralization'' sample results obtained with our method. 




Summarizing, we presented an exhaustive experimental evaluation showing several interesting results. First, the proposed S-cube approach effectively learns the seeds that are expected to produce the largest increase in memorability. This increase is consistently validated when measuring it with the external scorer $\mathbb{E}$. We also investigated the effect of the choice of architecture for the seed Selector and the effect of the amount of seeds in the overall performance. Finally, we have shown the per-image memorability increase when using the top few seeds, and varying the size of the seed set. In all, the manuscript provides experimental evidence that the proposed method is able to automatically increase the memorability of generic images.

\section{Conclusions}
This paper presented a novel approach to increase image memorability based on the \emph{editing-by-filtering} philosophy. Methodologically speaking, we propose to use three deep architecures as the Scorer, the Synthesizer and the Selector. The novelty of our approach relies on the fact that the Selector is able to rank the seeds according to the expected increase of memorability and select the best ones so as to feed the Synthesizer. The effectiveness of our approach both in increasing memorability and in selecting the top memoralizable style seeds has been evaluated on a public benchmark. 



We believe that the problem of increasing image memorability can have a direct impact in many fields like education, elderly care or user-generated data analysis. Indeed,   memorabilizing images could help editing educational supports, designing more effective brain training games for elderly people, producing better summaries from lifelog camera image streams or leisure picture albums.

While in this work we focused on memorability, the architecture of our approach of highly versatile 
and can potentially be applied to other concepts such as aesthetic judgement or emotions. A necessary condition to this is a sufficient precision of the Scorer, which should be as closer to human performance as possible. When this condition does not occur, the automatic prediction can be replaced introducing a data annotation campaign. 
The philosophy followed in this study could be extended to take into account other image properties such as aesthetics or evoked emotions \textit{simultaneously}. This is highly interesting and not straightforward, and we consider it as one of the main future work guidelines.

While literature on predicting image abstract concepts like memorability is quite huge, the literature in image synthesis with deep networks is still in its early infancy.
A promising line of work is represented by Generative Adversarial Networks (GANs) \cite{goodfellow2014generative}. 
However, it is not straightforward to apply GANs and still retaining the \emph{editing-by-filters} philosophy. Indeed, one prominent feature of our methodology is that we keep the user in the loop of the image manipulation process, 
by allowing them to participate to the style selection, once the most promising seeds are automatically provided. 
Future research works will also investigate an alternative holistic approach based on GANs.

%
\bibliographystyle{abbrv}
\bibliography{sigproc}  
%
%

\end{document}